\pdfoutput=1

\documentclass[11pt]{article}

\usepackage{naacl2021}

\usepackage{times}
\usepackage{latexsym}
\usepackage{todonotes}
\usepackage[clock]{ifsym}

\usepackage[T1]{fontenc}

\usepackage[utf8]{inputenc}

\usepackage{pifont}
\usepackage{microtype}

\usepackage{graphicx}

\title{Teaching NLP with Bracelets and Restaurant Menus: \\An Interactive Workshop for Italian Students}

\author{
  Ludovica Pannitto \\
   University of Trento \\
  \texttt{ludovica.pannitto@unitn.it} \\\And
  Lucia Busso \\
  Aston University \\
 \texttt{l.busso@aston.ac.uk} \\\AND
  Claudia Roberta Combei \\
  University of Bologna\\
 \texttt{claudiaroberta.combei@unibo.it~~~} \And
  Lucio Messina \\
  Independent Researcher \\
  \texttt{lucio.messina@autistici.org} \\
 \AND
 Alessio Miaschi \\
  University of Pisa\\
 \texttt{alessio.miaschi@phd.unipi.it~~~} \\\And
 Gabriele Sarti \\
 ~~~University of Trieste\\
 \texttt{~~~gsarti@sissa.it} \\\And
 Malvina Nissim \\
  University of Groningen\\
 \texttt{m.nissim@rug.nl} \\
 }

\begin{document}
\maketitle

\begin{abstract}
Although Natural Language Processing (NLP) is at the core of many tools young people use in their everyday life, high school curricula (in Italy) do not include any computational linguistics education. This lack of exposure makes the use of such tools less responsible than it could be and makes choosing computational linguistics as a university degree unlikely. To raise awareness, curiosity, and longer-term interest in young people, we have developed an interactive workshop designed to illustrate the basic principles of NLP and computational linguistics to high school Italian students aged between 13 and 18 years. The workshop takes the form of a game in which participants play the role of machines needing to solve some of the most common problems a computer faces in understanding language: from voice recognition to Markov chains to syntactic parsing. Participants are guided through the workshop with the help of instructors, who present the activities and explain core concepts from computational linguistics. The workshop was presented at numerous outlets in Italy between 2019 and 2021, both face-to-face and online. 
\end{abstract}

\section{Introduction}
\label{sec:intro}

\textit{Have you used Google this week?}
This  question would kick off the activity that we describe in this paper every time we delivered it. 
And a number of follow-up comments would generally appear. \textit{What for? Translating, getting some help for homework, looking for info, writing collaboratively - and getting spelling correction!}

In our workshops, we talk to groups of teenagers -- even if someone has not personally used any of those tools on a daily basis, it is  utmost unlikely that they have never interacted with a vocal assistant, wondered how their email spam filter works, used text predictions, or spoken to a chat-bot. Also, applications that do not require a proactive role of the user are growing: most of us, for example, are subject to targeted advertising, profiled on the content we produce and share on social media.

Natural Language Processing (NLP) has grown at an incredibly fast pace, and it is at the core of many of the tools we use every day.\footnote{In this discussion, and throughout the paper, we conflate the terms Natural Language Processing and Computational Linguistics and use them interchangeably.} At the same time, though, awareness of its underlying mechanisms and the scientific discussion that has led to such innovations, and even NLP's very existence as a scientific discipline is generally much less widespread and is basically unknown to the general public~\citep{grandi2018perche}.

A concurrent cause to this lack of awareness resides in the fact that in more traditional high-school formal education systems, such as the Italian one, ``young disciplines" such as Linguistics and Computer Science tend to be overlooked. 
Grammar, that in a high-school setting is the closest field to Linguistics, is rarely taught as a descriptive discipline; oftentimes, it is presented as a set of norms that one should follow in order to speak and write correctly in a given language. While this approach has its benefits, it is particularly misleading when it comes to what actual linguistic research is about. Similarly, Computer Science is often misread by the general public as an activity that deals with computers, while aspects concerning information technology and language processing are often neglected.
This often leads to two important consequences. First, despite the overwhelming amount of NLP applications, students and citizens at large lack the basic notions that would allow them to fully understand technology and interact with it in a responsible and critical way. Second, high-school students might not be  aware of Computational Linguistics as an option for their university degree. Oftentimes, students that enrol in Humanities degrees are mainly interested in literature and they only get acquainted with linguistics as discipline at university. 
At the same time, most Computer Science curricula in Italian higher education rarely focus on natural language-based applications. As a result, Computational Linguistics as such is practically never taught before graduate studies.

As members of the Italian Association for Computational Linguistics (AILC, \url{www.ai-lc.it}) we have long felt the need to bridge this knowledge gap, and made dissemination a core goal of the Association. As a step in this direction, we have developed a dissemination activity that covers the basic aspects of what it means to process and analyze language computationally. This is the first activity of its kind developed and promoted by AILC, and to the best of our knowledge, among the first in Italy at large.

This contribution describes the activity itself, the way it was implemented as a workshop for high school students in the context of several dissemination events, and how it can serve as a blueprint to develop similar activities for yet new languages.

\section{Genesis and Goals}

We set to develop an activity whose main aim would be to  provide a broad overview of language modeling, and, most importantly, to highlight the open challenges in language understanding and generation. 

Without any ambition to present and explain the actual NLP techniques to students, we rather focused on showing how language, which is usually conceptualized by the layperson as a simple and monolithic object, is instead a complex stratification of interconnected layers that need to be disentangled in order to provide a suitable formalization.

In developing our activity, we took inspiration from the \textit{word salad} Linguistic Puzzle, as published in \newcite{radev2013puzzles}:
\begingroup
\addtolength\leftmargini{-0.1in}
\begin{quote}
Charlie  and  Jane  had  been  passing  notes  in  class,  when  suddenly  their  teacher  Mr.  Johnson  saw  what  was  going  on. He  rushed  to  the  back  of  the  class,  took  the  note  Charlie  had  just  passed  Jane,  and  ripped  it  up,  dropping  the  pieces  on  the  floor. Jane  noticed  that  he  had  managed  to  rip  each  word  of  the  message  onto  a  separate  piece  of  paper.    The  pieces  of  paper  were,  in  alphabetical  order,  as  follows: \textit{dog, in, is, my, school, the}. Most  likely,  what  did  Charlie's  note  originally  say?
\end{quote}
\endgroup

The problem includes a number of follow up questions that encourage the student to reflect upon the boundaries of sentence structure.
In particular, we found that the \textit{word salad} puzzle would give us the opportunity to introduce some of the core aspects of Computational Linguistics' research. Approaching the problem with no previous knowledge helps raising some crucial questions, such as: \textit{what are the building blocks of our linguistic ability that allow us to perform such a task?}, \textit{how much knowledge can we extract from text alone?}, \textit{what does linguistic knowledge look like?}

Since the workshop here presented is the first activity of this kind in the Italian context, we took inspiration from games and problems such as those outlined in \newcite{radev2013puzzles} and used for the North American Computational Linguistics Olympiads, similar to the ones described in \newcite{van2002teaching} and \newcite{iomdin2013linguistic}. Particularly, we were inspired by the institution of (Computational) Linguistic Olympiads in making our workshop a problem-solving game with different activities, each related to a different aspect of computational language processing.
Linguistic Olympiads are now an established annual event in many parts of the world since they first took place in Moscow in 1965. In these competitions students (generally of high-school age) are faced with linguistic problems of varying nature, that require participants to use problem-solving abilities to uncover underlying patterns or rules in the data. For an in-depth discussion of the history and diffusion of Linguistic Olympiads in the world, see \newcite{derzhanski2010} and \newcite{littell2013}.
\begin{table*}[ht!]
\center{
\begin{tabular}{lcp{8cm}}
\hline
\multicolumn{1}{c}{\textbf{Activity}}               &
\StopWatchStart &
\multicolumn{1}{c}{\textbf{Aim}} \\
\hline
1. Get to know a (computational) linguist  & 10'   & collaboratively build a definition for \textit{linguistics} as a study field  \\ \hline
2. Are computers able to \textit{hear}?    & 15'   & familiarize with the concept of \textit{simulation} of humans' speech perception abilities \\ \hline
3. Are computers able to \textit{read}?    & 30'   & introduce corpora as sources of linguistic knowledge and statistical patterns as structural aspects of language \\ \hline
4. Do computers \textit{know} grammar?     & 30'   & introduce human annotation and meta-linguistic knowledge as powerful research tools \\ \hline
5. Becoming a computational linguist       & 5'    & evaluate pros and cons of the two presented approaches, future directions and discuss about what's needed to become a computational linguist \\
\hline
\end{tabular}
}
\caption{Sections of the activity, with their planned duration and a broad aim for each of them.}
\label{tab:table-times}
\end{table*}

In the choice of algorithms to include in our dissemination activity, we decided to leave aside neural networks and instead focus on traditional statistical approaches, both for historical reasons and for the fact that these convey more clearly the distinction between different layers of linguistic information and their roles in language modeling. 
A brief discussion of most recent NLP technologies, including the application of neural networks, is included in the final part of the workshop (Sec.~ \ref{becoming_a_comp_ling}).

The activity is targeted at students in their last year of middle school (13 years of age) or older. While we believe 13 is a good entry point, there isn't an actual upper-bound, since the activity can be enjoyed by people of any older age (though in practice participation was mainly offered to schools, with the oldest students being 18-19).  
We thought this would be the appropriate target audience of this workshop for two main reasons. On the one hand, we believe that coming to the activity with a richer metalinguistic background, typically acquired during the first Italian school cycle, would be beneficial for the attendees to better grasp the differences between the scientific approach to language and the more prescriptive approach they are exposed to in school. On the other hand we also conceived the activity as a way of helping students in their future study choices: we therefore included both students attending their last year of middle school and therefore about to choose a high school curriculum as well as high school students, the latter in order to provide them with more options for their university choices. 

\section{The activity}
\label{sec:activity}

We planned our dissemination activity for a 90 minutes time slot, divided into five parts, as detailed in Table~\ref{tab:table-times}.

\subsection{Get to know a (computational) linguist}

The first 15 minutes of the workshop are organized both as an ice-breaker activity for the attendees, and as a brief introduction to linguistics and computational linguistics more specifically.

During the introduction we tried to demystify some common misconceptions about linguistics, (i.e., \textit{a linguist knows many languages}, \textit{linguists get sometimes confused with speech therapists}, \textit{a linguist will correct my grammar}, etc.): we presented participants with a list of possible definitions, and they had to identify appropriate ones. 
We broadly defined linguistics as the study of language as a biological, psychological, and cultural object. Computational Linguistics was then introduced both as a commercial and engineering-oriented field, as well as a purely scientific research discipline.\footnote{We relied on the definition reported in \url{https://www.thebritishacademy.ac.uk/blog/what-computational-linguistics/}.}

We also briefly discussed \textit{linguistic questions} with participants as an example of the kind of problems that a linguist tries to approach during their research activity. These included: "How many ways of pronouncing \textit{n} do we have in Italian?", "Do numbers from one to ten exist in all languages?".

For the following parts of the activity, the introduction to each sub-part was dedicated to a reflection upon what it means for us humans to \textit{hear}, \textit{read} and \textit{understand grammar}, and whether there is a difference when the same tasks are performed by computers.

\subsection{Are computers able to \textit{hear}?}
\label{sec:hear}

As vocal assistants such as Alexa, Siri, Google Home etc. have become  increasingly popular, we chose them as tangible examples of NLP technologies to kick-off the games. The aim of this section of the workshop was to demonstrate two points:
\begin{itemize}
\itemsep 0pt
    \item computers do not necessarily solve linguistic tasks the way we solve them; they are therefore \textit{simulating} our abilities without \textit{replicating} them;
    \item consequently, the concepts of \textit{easy} and \textit{difficult} tasks have to be carefully revised when applied to language models.
\end{itemize}

\begin{figure}
    \centering
    \includegraphics[width=0.5\textwidth]{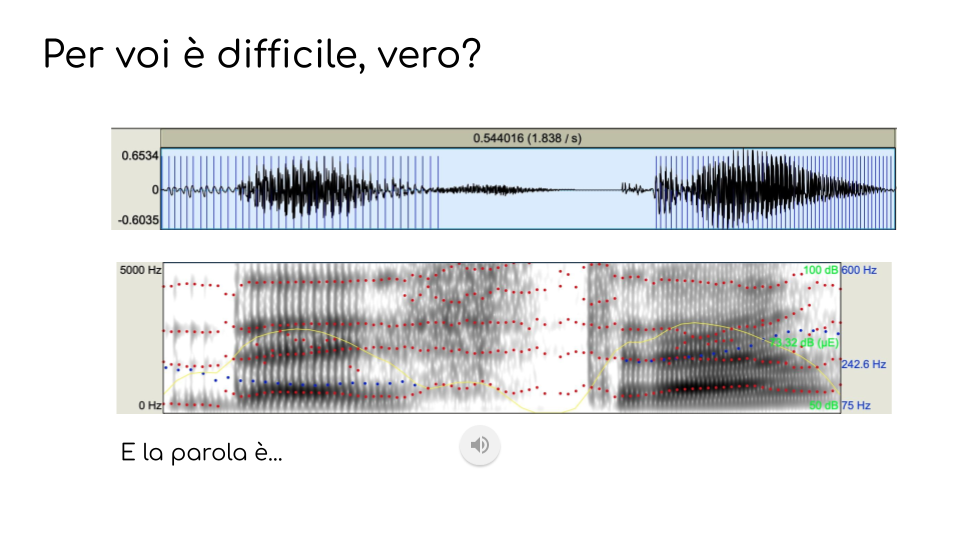}
    \caption{Representation of the audio signal for the italian word \textit{destra} (en. \textit{right}). This was used to show how information coming from audio signals can be represented in a way that easily allows the computer to perform a pattern matching task, but would be impossible to process for  humans.}
    \label{fig:spettrogramma}
\end{figure}

\noindent We introduced the McGurk effect \citep{mcgurk1976hearing}, to clarify how hearing language is a complex task in itself, involving a broad set of aspects beyond simple sound perception, such as the visual system as well as the expectations regarding the upcoming input. Computers on the other hand \textit{hear} on the basis of an audio signal that is processed (Figure~\ref{fig:spettrogramma}), at least in the most traditional and well-established architectures, without access to higher level linguistic knowledge or information from the communicative context.

We then briefly presented speech recognition as a direct optimization task (i.e., given an audio signal, find the word in a given database that maximises the probability of being associated to that signal) and introduced one of the major challenges that speech recognition models are still facing, namely the ability to adapt to different speech styles (i.e., speakers of different language varieties and dialects, non-native speakers, speakers with impairments etc.).

In order to further demonstrate this, we tested the attendees' ability to adapt their hearing skills to different speech varieties by making them hear conversations in various Italian regional accents.\footnote{Conversations were extracted from corpus CLIPS \citep{albano2007clips}.} 
While we asked attendees to guess the name of the region of the speakers, the actual aim was to show how we easily adapt to understand speech, differently from speech recognition systems.

\subsection{Are computers able to \textit{read}?}
\label{sec:read}

From this moment on, the attendees worked on written text. The activity described in this section was aimed at showing how salient aspects related to language structure can be derived from the statistical properties of language.

\begin{figure}[htb]
    \centering
    \includegraphics[width=0.4\textwidth]{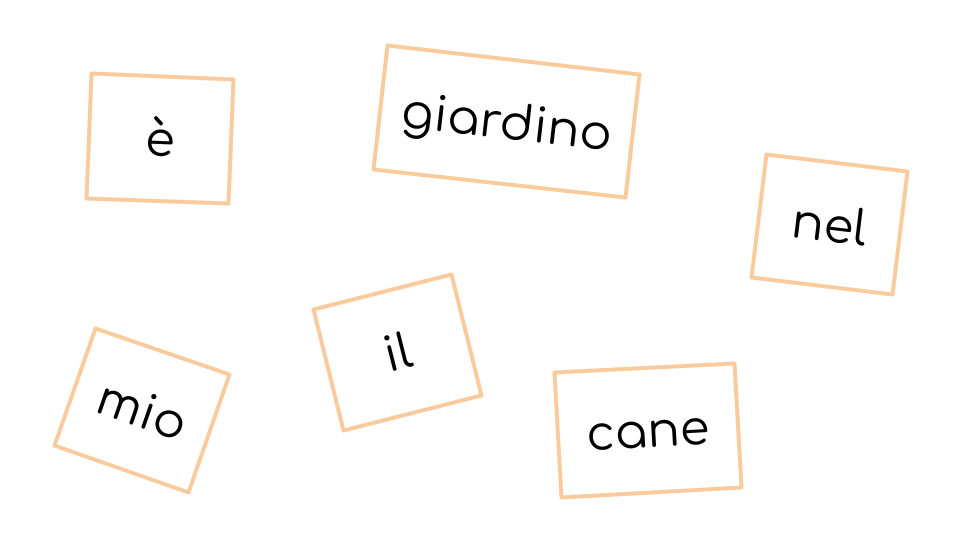}
    \caption{A set of Italian words (from the top left corner, en. \textit{is, garden, in, my, the, dog}): when asked to re-arrange them into a sentence, participants would first come up with the most likely ordering (i.e., \textit{il mio cane è nel giardino}, en. \textit{my dog is in the garden}) and if prompted they would then produce more creative sentences (i.e., \textit{il cane nel giardino è mio}, en. \textit{the dog in the garden is mine}). They would however never consider ungrammatical sequences  as possible sentences.}
    \label{fig:parole-ita}
\end{figure}

\begin{figure}[htb]
    \centering
    \includegraphics[width=0.4\textwidth]{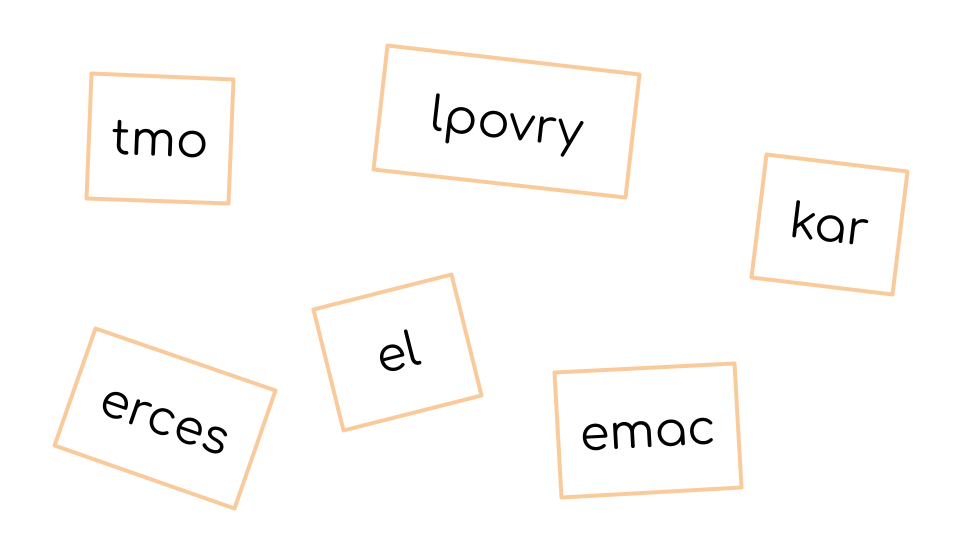}
    \caption{The figure depicts the same situation as Figure~\ref{fig:parole-ita}, this time with non-words.}
    \label{fig:parole-mask}
\end{figure}

Following the ``Word Salad'' puzzle~\citep{radev2013puzzles}, the ability of \textit{reading} was presented as follows: given a set of words, are we able to rearrange them in a plausible sentence-like order?
We demonstrated how this is an easy task for human beings, when one deals with a language they are familiar with (Figure~\ref{fig:parole-ita}), while, generally one may not be able to perform the same task in case of unknown languages (Figure~\ref{fig:parole-mask}), where each possible ordering seems equally plausible.

\begin{figure}
    \centering
    \includegraphics[width=0.4\textwidth]{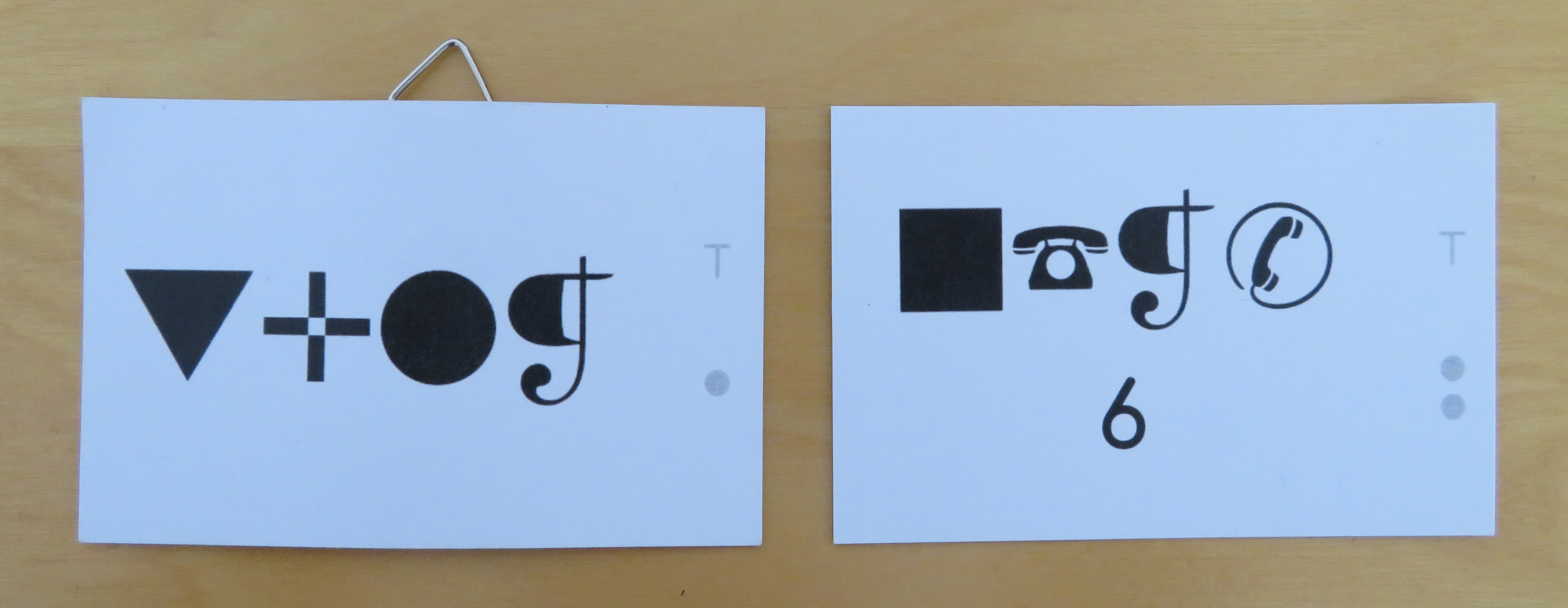}
    \caption{Example cards, both showing a word. Left: a card for the first activity, with a button loop to thread it in a sentence. Right: a card for the second activity, with part of speech (number) at the bottom.}
    \label{fig:tokens-1-2}
\end{figure}

We therefore gave the attendees a deck of unknown, mysterious tokens (left card in Figure~\ref{fig:tokens-1-2}) and asked them to come up with the most plausible sentence that contained all of them. The cards represented either Italian or English words (participants were divided into two teams, each one dealing with a different \textit{masked} language) which had however been transliterated into an unknown alphabet.
While this was obviously an impossible task to solve, it gave us the opportunity to introduce a notion of probability in the linguistic realm. We cast it as the expectation that we humans bear on the appearance of a specific linguistic sequence, and the subsequent need for a source of linguistic knowledge to implement the same notion.

When asked to perform the same task on the words of Figure~\ref{fig:parole-ita}, participants produced sentences in a quite consistent order, and the most prototypical sentences (e.g., \textit{il mio cane è nel giardino}, en. \textit{my dog is in the garden}) were usually elicited before some less typical ones (e.g., \textit{il cane nel giardino è mio}, en. \textit{the dog in the garden is mine}), while agrammatical sentences (i.e., random permutations) were never produced.

We justified their responses by explaining that humans accumulate a great amount of linguistic knowledge throughout their lifetime that helps them refine these expectations, while machines are instead in principle unbiased towards having any specific preference. This observation allowed us to introduce participants to the notion of \textit{corpus} as a large collection of linguistic data that mimics the amount of data we are exposed to as humans.

\begin{figure}
    \centering
    \includegraphics[width=0.45\textwidth]{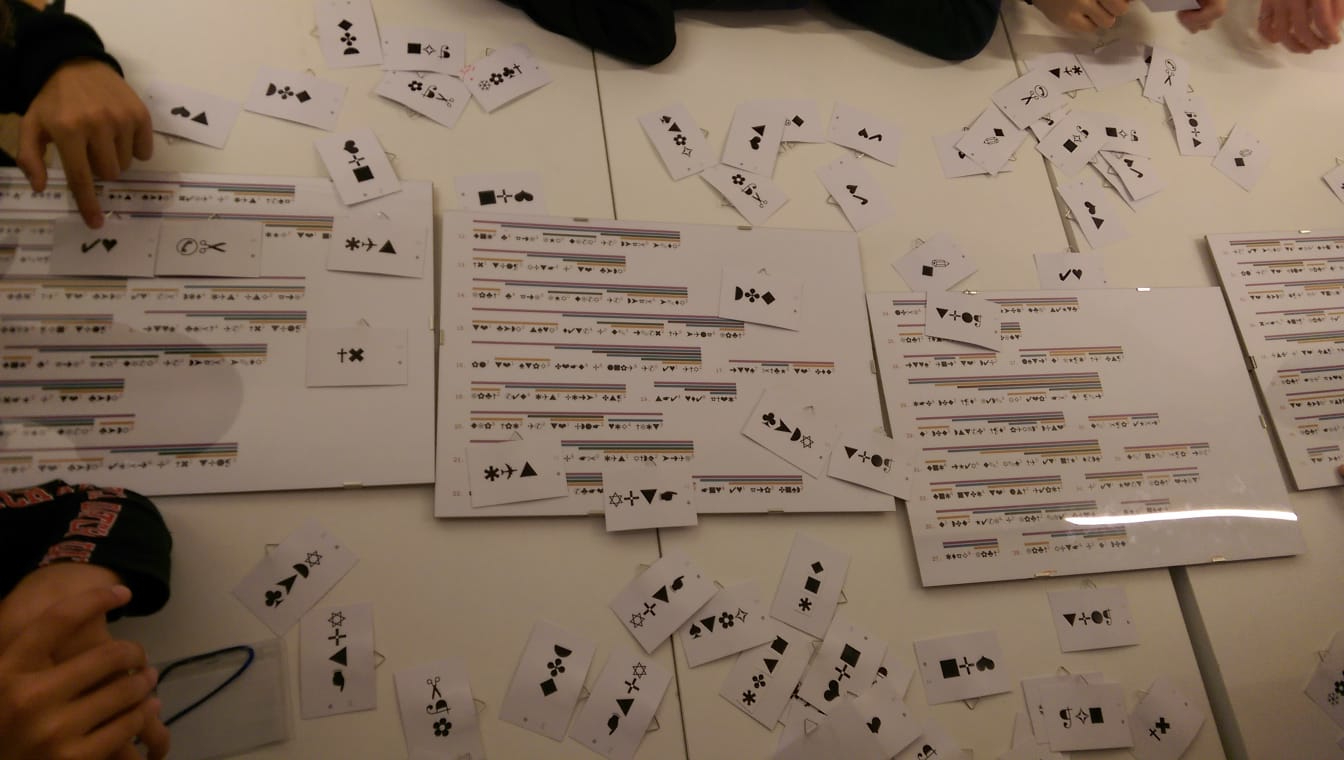}
    \caption{The figure shows one of the corpora that was given to participants, 5 A3 tables containing approx. 60 transliterated sentences from the ``Snow White'' fairy tale, and tokens with buttonholes that had to be searched in the corpus and threaded into sentences.}
    \label{fig:corpus-bergamo}
\end{figure}

Each team was then provided with a corpus written in an unknown language (approximately 60 sentences hand-crafted by transliterating a portion of the ``Snow White'' tale into a mysterious alphabet Figure~\ref{fig:corpus-bergamo}). Concurrently, we introduced a simple algorithm to tell apart sentence-like orderings of the provided tokens from the random ones.

\begin{figure}[t]
    \centering
    \includegraphics[width=0.5\textwidth]{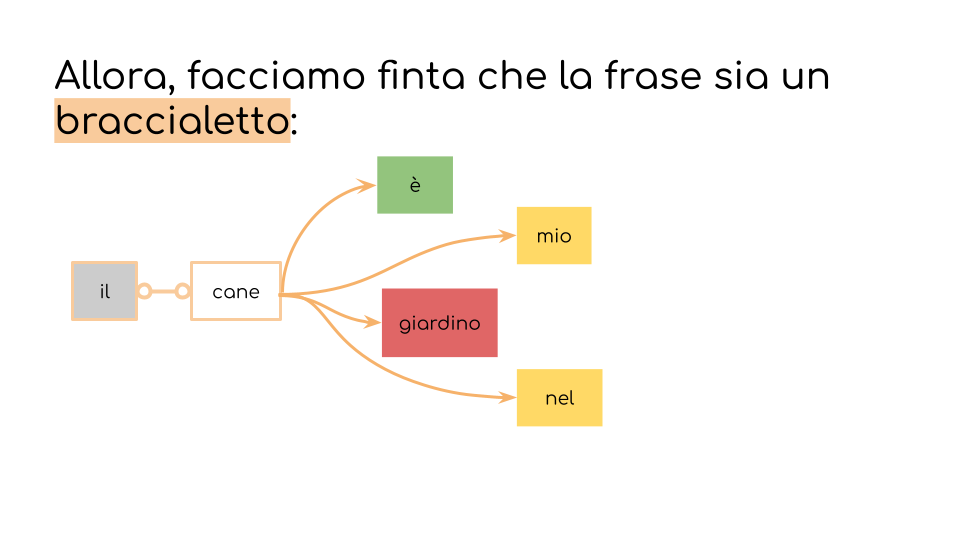}
    \caption{The \textit{bracelet method} applied on the Italian words \textit{è, mio, giardino, nel} (en. \textit{is, my, garden, in}). The algorithm is based on bigram co-occurrences, so the choice for each word is based solely on the previously chosen one. Colors indicate probabilities, which are computed based only on the previously chosen word \textit{cane} (en. \textit{dog}). The first token, \textit{il} (en. \textit{the}), appears grayed as it is ignored for the choice.} 
    \label{fig:braccialetto}
\end{figure}

The algorithm, which we called \textit{The bracelet method} (Figure~\ref{fig:braccialetto}), is based on the Markov Chain procedure: in a scenario similar to that of lining pearls up to form a bracelet, participants could decompose the task of forming up a sentence into smaller tasks. 
To make the operation more concrete, we equipped each card with a button loop as shown in Figure~\ref{fig:tokens-1-2}: this way cards could be physically threaded together to form a sentence.

The first step consisted in choosing the first word, for the beginning of the sentence. Since participants were facing a language they didn't know, they  were not aware of language structures nor of the meaning of the tokens. In such a situation, they could decide whether it was plausible to use a given word at the beginning of a sentence just by looking up in the corpus sentences that began the same way. If they found at least one sentence that began with the same word, it meant that that was a licit position and they could use it to start the sentence.

The activity continued as follows: sticking to the bracelet metaphor, participants needed to select and insert the following "pearl" into the thread: ideally the pearl should pair well with the previous one, as we might want to avoid colour mismatch (e.g., it is well known that blue and green do not go well together).
The metaphor highlights therefore a core aspect that holds true for language as well, namely that we can condition our choice on a variable number of previous choices, and this will influence the complexity of the obtained pattern.

\begin{figure}[t]
    \centering
    \includegraphics[width=0.4\textwidth]{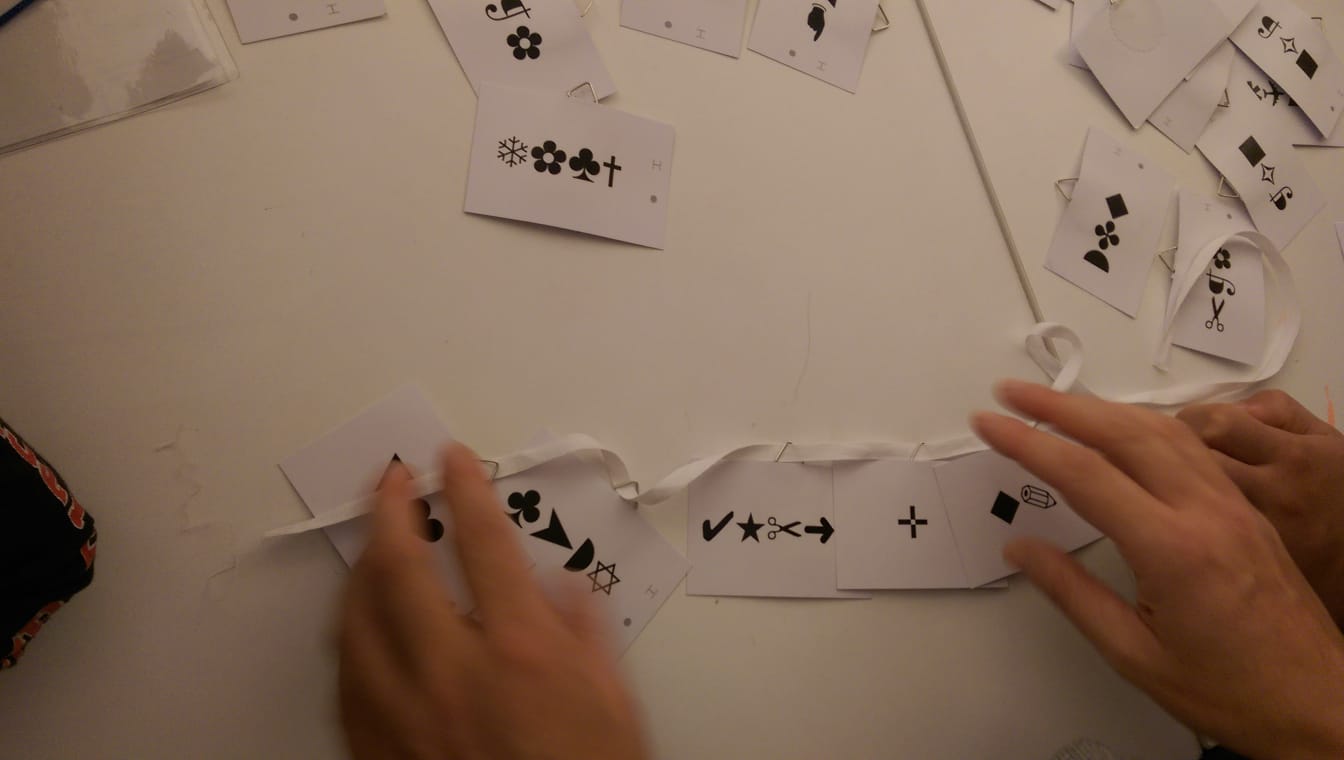}
    \caption{The figure shows the result of a bracelet sequence: tokens are threaded together based on co-occurrences in the corpus.}
    \label{fig:bracelet-bergamo}
\end{figure}

The activity resulted in a number of sentence-bracelets, as shown in Figure~\ref{fig:bracelet-bergamo}, that were then kept aside to be translated at the end of the workshop.

\subsection{Do computers \textit{know} grammar?}
\label{sec:grammar}

While the previous game highlighted the importance of statistical information in NLP, in accordance with the overall aim of the activity, we also wanted to introduce some of the algorithms that are more deeply rooted in the linguistic tradition.

\begin{figure}[!h]
    \centering
    \includegraphics[width=0.5\textwidth]{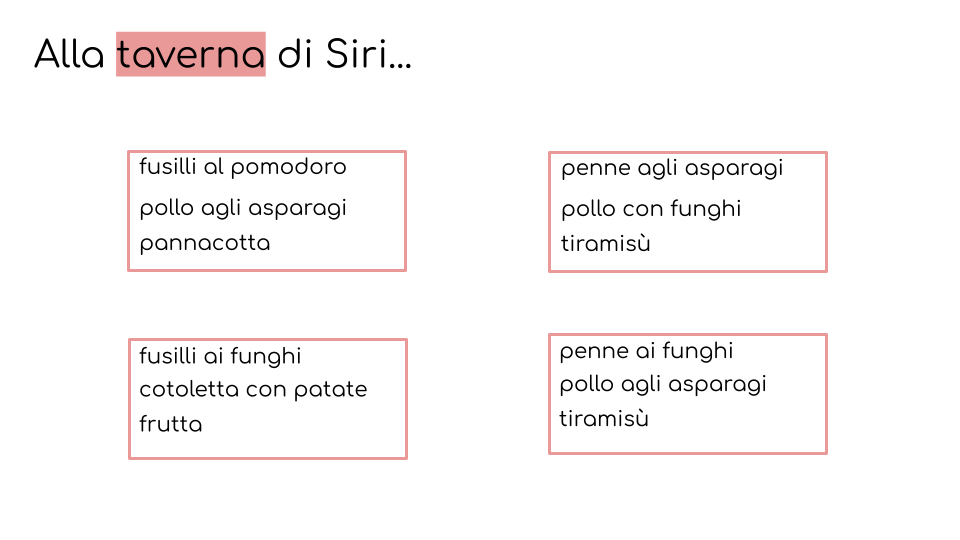}
    \caption{Each block in the image corresponds to a possible Italian full meal, consisting of: first course (e.g. \textit{fusilli al pomodoro}), second course (e.g. \textit{pollo agli asparagi}) and dessert (e.g. \textit{pannacotta}).}
    \label{fig:menu}
\end{figure}

\begin{figure}[!h]
    \centering
    \includegraphics[width=0.5\textwidth]{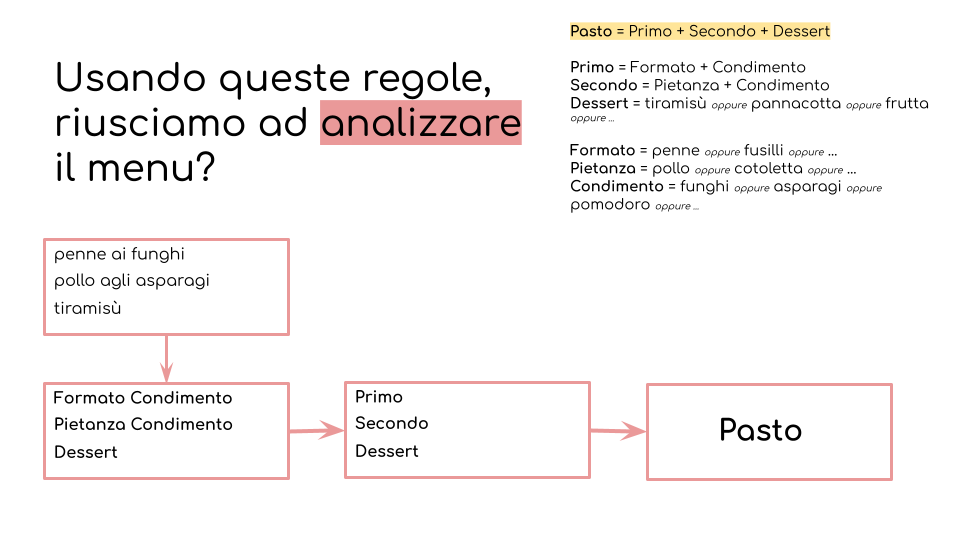}
    \caption{Step-by-step process to assess the validity of a given menu. In the top-right corner the rules for creating a full meal are shown . Categories are defined recursively until each course that constitute the full menu is obtained and therefore reduced to the initial category of a \textit{pasto} (en. \textit{meal}).}
    \label{fig:riduzione}
\end{figure}

\begin{figure}[!h]
    \centering
    \includegraphics[width=0.5\textwidth]{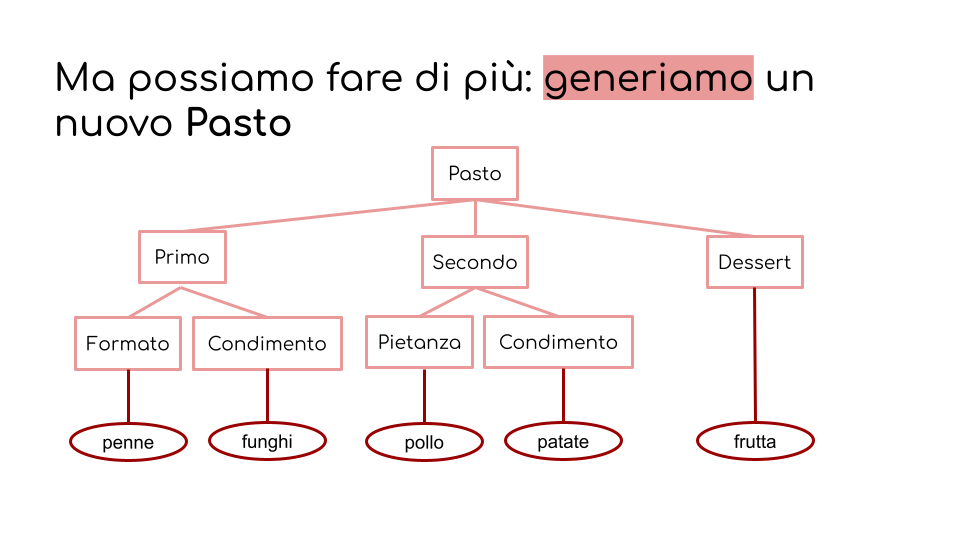}
    \caption{Process flow for creating a new meal from the initial category \textit{pasto} (en. \textit{meal}) up to the leaves containing terminal symbols such as \textit{penne}, \textit{funghi}, \textit{pollo} etc. (en., a type of \textit{pasta}, \textit{mushrooms}, \textit{chicken}). The rules used to generate are the same used for the reduction process, reported in Figure~\ref{fig:riduzione}.}
    \label{fig:generazione}
\end{figure}

In order to do so, we introduced the notion of grammar as a descriptive abstraction over a set of examples. Out of the linguistic context, to exemplify this notion of grammar metaphorically, we presented participants with a set of possible restaurant menus (Figure~\ref{fig:menu}), and encouraged them to come up with the general rule that the restaurant owner must have had in mind when choosing those combinations. All menus were built as a traditional Italian full meal, composed by two main dishes and a dessert.
We perpetuated the metaphor showing how, once a set of rules is defined, these can be used both to decide if a new menu is likely to be part of the same set (Figure~\ref{fig:riduzione}) and also to generate new meals   (Figure~\ref{fig:generazione}).

This metaphor, which was readily grasped by most participants, was useful to show how different components can be combined together in a meaningful way, as it happens in grammar. Before moving back to the corpus, we showed them what a formal grammar of the menus could look like.

\begin{figure}[t!]
    \centering
    \includegraphics[width=0.5\textwidth]{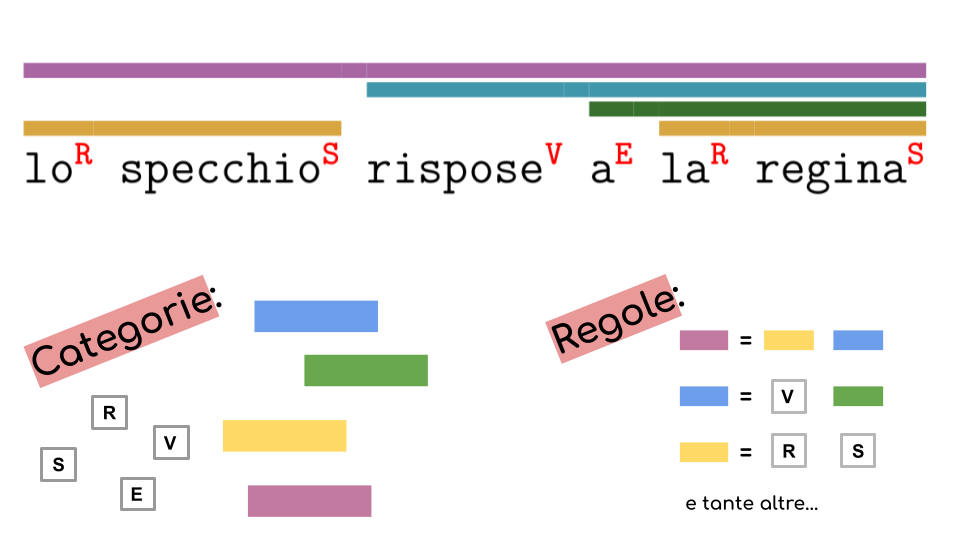}
    \caption{Example of categories (it. \textit{Categorie}), e.g. phrases and POS tags, and rules (it. \textit{Regole}) for the sentence \textit{"lo specchio rispose a la regina"} (en. \textit{the mirror answered to the queen}).}
    \label{fig:corpus-annotato}
\end{figure}

We had previously annotated the corpus with syntactic and morpho-syntactic information (i.e., part of speech), as shown in Figures~\ref{fig:corpus-bergamo} and \ref{fig:corpus-annotato}.
We therefore asked participants to extract from the corpus a possible grammar, namely a set of attested rules and use them to generate a new sentence. 

\begin{figure}[!h]
    \centering
    \includegraphics[width=0.38\textwidth]{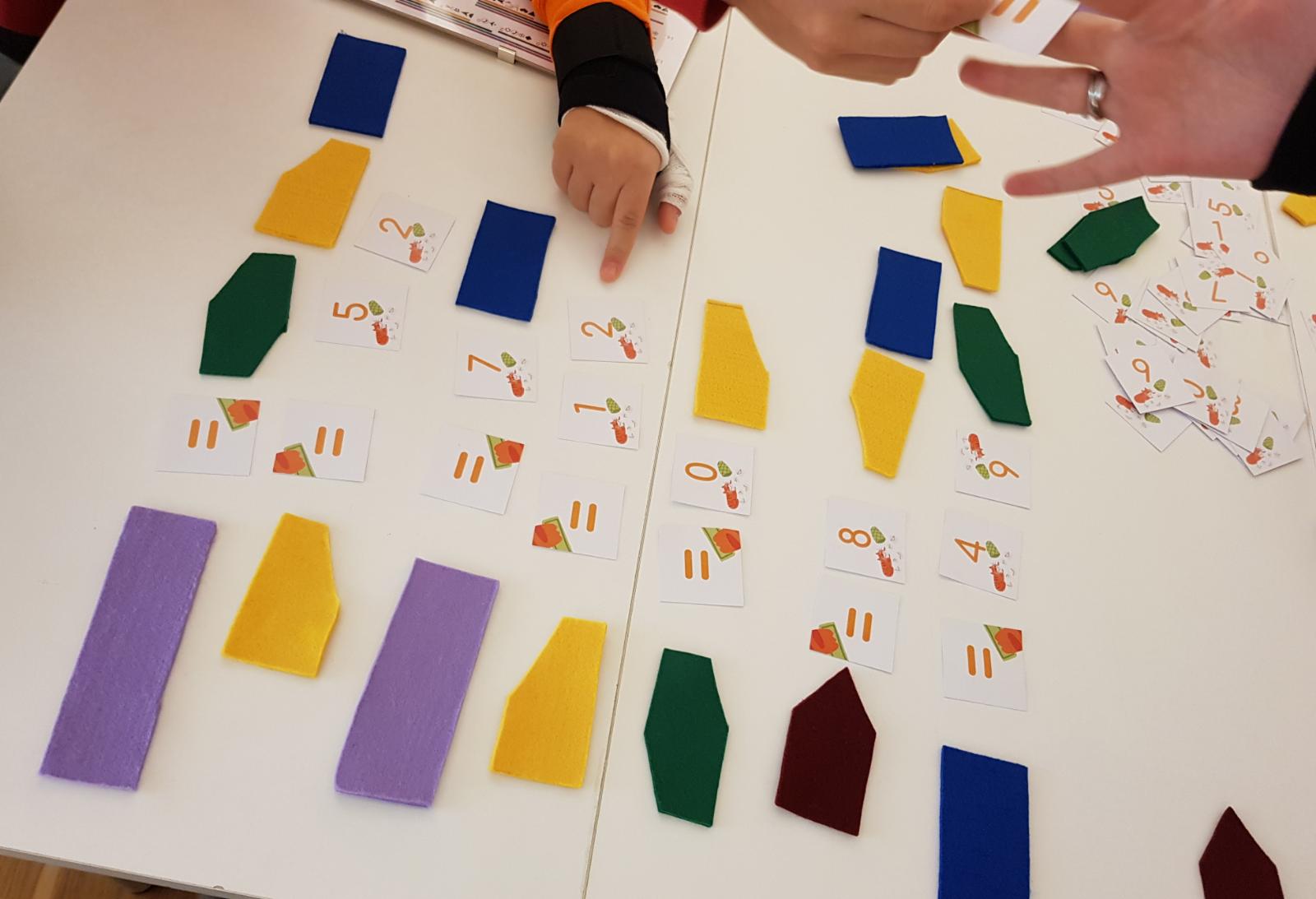}
    \caption{A set of rules extracted during the activity from the corpus. Each rule is made of felt strips for phrases, cards with numbers indicating parts of speech, and ``$=$'' cards.}
    \label{fig:rules}
\end{figure}

In order to write the grammar, participants were given several physical materials: felt strips reproducing the colors of the annotation, a deck of cards with numbers (identifying parts of speech) and a deck of ``$=$'' symbols (Figure~\ref{fig:rules}).

With a new deck of words (Figure~\ref{fig:tokens-1-2}, right panel), not all of which present in the corpus, participants had to generate a sentence using the previously composed rules.

\subsection{Becoming a computational linguist}
\label{becoming_a_comp_ling}

\begin{figure}
    \centering
    \includegraphics[width=0.4\textwidth]{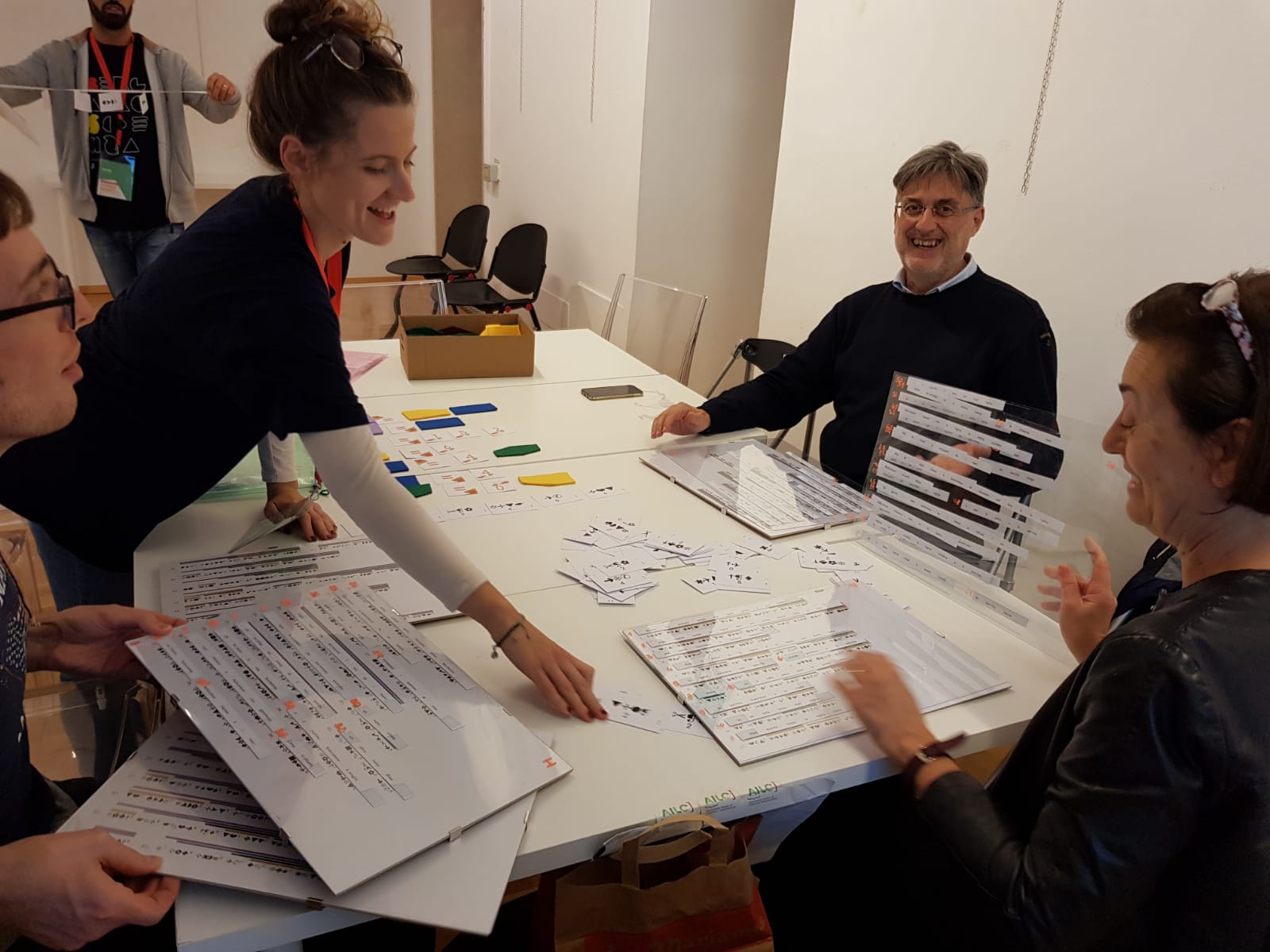}
    \caption{A trial session of the workshop (the picture shows some of the tutors explaining the game to AILC members): the original language of the corpus has just been revealed by superimposing Plexiglas supports on the corpus tables.}
    \label{fig:plexiglass}
\end{figure}

At this point, participants had created a number of sentences by means of the two techniques described above.
It is now time to discover that the mysterious languages they worked on were actually English and Italian. 
This was achieved in practice by superimposing a Plexiglas frame on the A3 corpus pages (Figure~\ref{fig:plexiglass}): the true nature of the corpora was this way revealed as the participants could see the original texts (in Italian and English) and translate the sentences they had created previously.

The outcome of the activity stimulated discussion amongst the participants, highlighting pros and cons of each approach and how could they be integrated into real-life technologies.
Our workshop ended with a brief description of more recent NLP technologies and their commercial applications, such as recommender systems, automatic translation, text completion,  etc. 

Since the target audience consisted mostly of middle- and high-school students, we offered an overview of what it takes to become a computational linguist and where one could study computational linguistics in Italy.

\section{The workshop in action}

The activity -- here outlined in its complete and original form -- was presented in various outlets during the last year and a half.

It was initially designed for the 2019 edition of the ``Bergamo Scienza'' Science Festival\footnote{\url{https://festival.bergamoscienza.it/}}, where it was presented live to over 450 participants in the course of two weeks. A simplified "print-and-play" version of the workshop was also presented at the 2020 edition of the SISSA\footnote{\url{https://www.sissa.it/}} Student Day.

Due to the Covid-19 pandemic, all other presentations of the activity had to be moved online. Transposing the workshop crucially meant striving to maintaining the interactive nature of the activities without the possibility of meeting face to face. To do so, we integrated our original presentation on Google slides with the interactive presentation software Mentimeter\footnote{\url{https://www.menti.com/}} - which we used for questions, polling and quizzes. The corpus and tokens were presented via a web interface created especially for this purpose\footnote{A demo of the interface can be found at \url{https://donde.altervista.org/}}.

\begin{figure*}
    \centering
        \includegraphics[width=\textwidth]{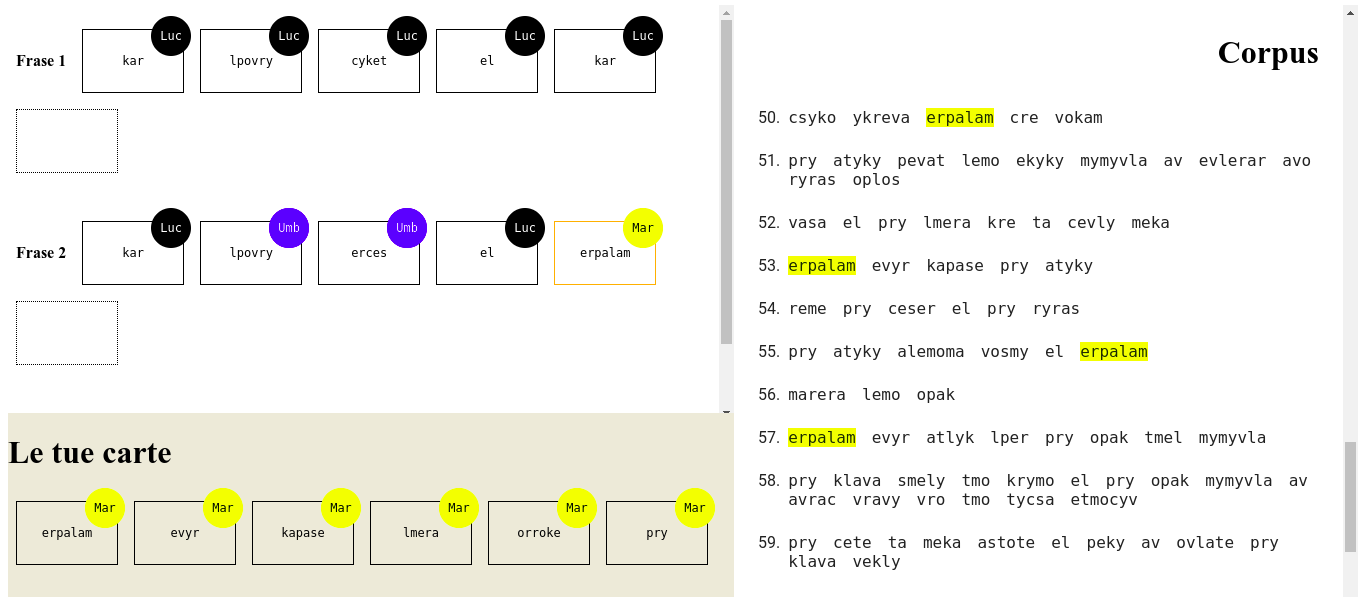}
    \caption{Interface of the online website used for the "bracelet" game (section \ref{sec:read}) during the online workshops. Players can collaboratively drag and drop their card from the bottom left panel to form sentences in the top-left panel. The corpus is shown in the right panel.}
    \label{fig:piattaforma-1}
\end{figure*}

\begin{figure*}[t]
    \centering
    \includegraphics[width=\textwidth]{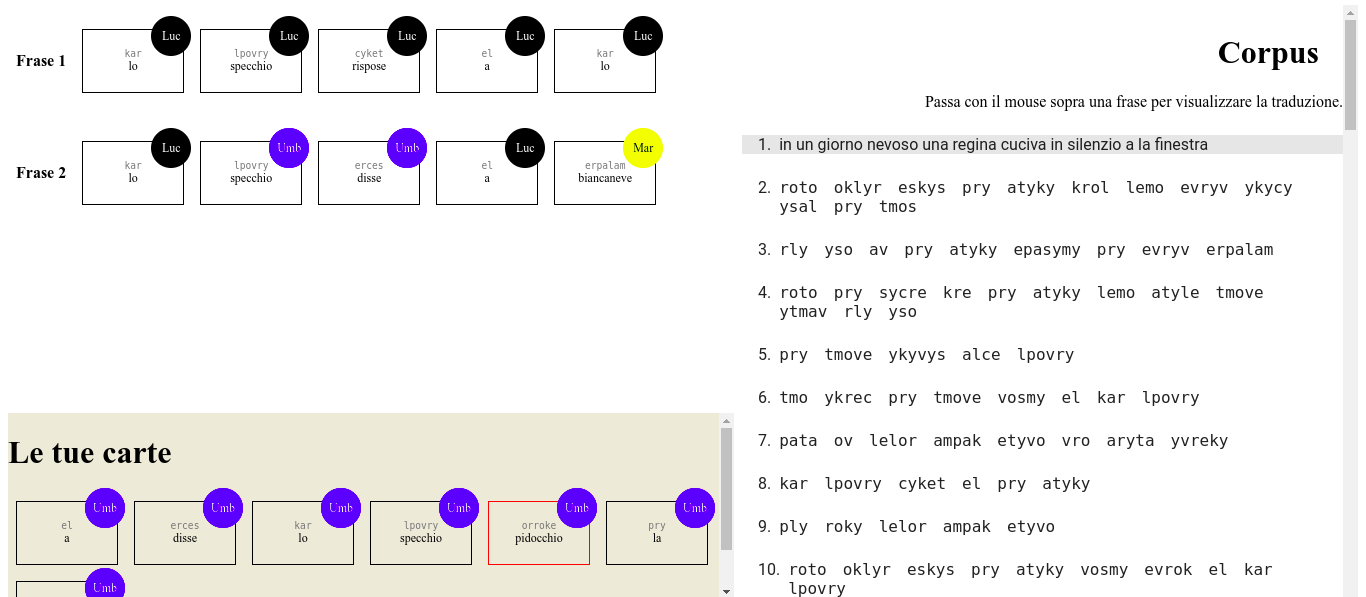}
    \caption{After the games the website shows the translation of the corpus (see for example line~1) and cards.}
    \label{fig:piattaforma-2}
\end{figure*}

The interface presented the masked corpus, complete with POS tags and syntactic annotations. For the bracelet activity participants were automatically assigned a number of tokens which they could use to build a sentence by simply dragging and dropping them. (Figures~\ref{fig:piattaforma-1}~and~ \ref{fig:piattaforma-2}).

\begin{figure*}[h]
    \centering
    \includegraphics[width=\textwidth]{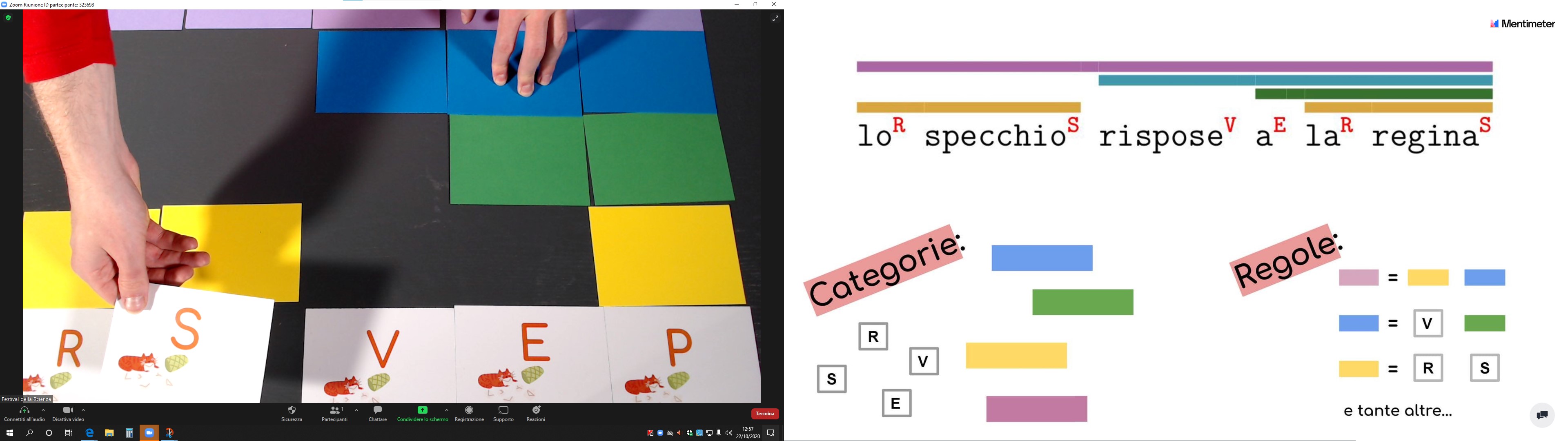}
    \caption{The picture was taken during a workshop session at ``Festival della Scienza'' in Genoa, in October 2020. Due to the COVID-19 pandemic, students were participating remotely. The left panel shows the tutor handling grammatical categories (colored cards for syntactic phrases and letters for parts of speech); the right panel shows a screenshot of the slides employed during the activity. Students could see both panels as slides were streamed while a webcam was capturing the tutor's movements.}
    \label{fig:genova}
\end{figure*}

This online version was crafted in the first place to be presented at ``Festival della Scienza''\footnote{\url{http://www.festivalscienza.it}} (Figure~\ref{fig:genova}), a science festival primarily aimed at school students held each year in Genoa, where multiple 45' sessions of the workshop were run over the course of two days.
The fourth activity (Section~\ref{sec:grammar}) involved "bootstrapping" syntactic rules based only on our color-based annotation. To simplify online interaction, we only used the unmasked Italian version of the corpus and participants played collectively, helping each other to build sentences and grammatical rules: rules were collected through a Mentimeter poll, while a sentence was generated in a guided demonstration by the presenter of the workshop.

Overall, the workshop transposed really well online, and was extremely well-received in this version as well. The online modality also allowed us to present it to a more vast and varied audience than just students: a version for the general public was presented at \textit{European Researchers' Night} (\textit{Bright Night}\footnote{\url{https://www.bright-night.it/}}) at the University of Pisa, thanks to a collaboration with the Computational Linguistics Laboratory\footnote{\url{http://colinglab.humnet.unipi.it/}} and ILC-CNR\footnote{\url{http://www.ilc.cnr.it/}} in November 2020, a dedicated session was run for the High school ITS Buzzi\footnote{\url{https://www.tulliobuzzi.edu.it/}} (Prato, Italy) in December 2020, and during the second edition of the Science Web Festival\footnote{\url{https://www.sciencewebfestival.it/}} in April 2021.

\section{Reusability: this activity as a blueprint}

As mentioned in Section~\ref{sec:intro}, our activity was inspired by  a collection of English-language problems created for students participating in the Computational Linguistics Olympiad \citep{radev2013puzzles}.

We adapted the original activity to the Italian language and context. While we tried to choose widely shared linguistic principles to communicate, the operation of adapting the game to a different language obviously requires language specific choices and details, which would have to be re-evaluated when porting the activity to yet new languages.  
Particularly, since the materials have been developed for Italian, it might be the case that transposition is not straightforward for some languages. However, we believe that the general structure of our workshop can serve as a blueprint for extension to new languages. For this reason all the relevant materials are made available in a dedicated repository. The repository includes both scripts to reproduce our activity as well as a general set of insights/recommendations regarding the structure and principles of the workshop.

All of the scripts necessary to produce the materials used in the game's workflow in a different language are made available in our open-access repository\footnote{\url{https://bitbucket.org/melfnt/malvisindi}}. To get the activity into production using the scripts provided, one only needs to create an annotated corpus in the target language. 

Our specific choice of ``Snow White'' as a corpus is motivated by the fact that the story can be phrased in a fairly repetitive formulation, with two advantages. One is that enough bigrams are present that enable the generation of new sentences with the \textit{bracelet method}. The other one is that the story contains recognizable characters (e.g., the dwarfs, the evil queen etc.), so that, when the unmasked text is revealed, the process results intuitively transparent. Such characteristics are desirable for the activity, and should be kept in mind when choosing a new text for a new language.

In addition to sharing scripts, we are sharing here the core structure and principles the workshop relies on, which can be reproduced when replicating the activity also for a different language.

\paragraph{Parts~1\&2}
At the beginning of the workshop (see Section~\ref{sec:hear}) we show the limits of current technologies, in particular in terms of adaptability. To this end, we exploited diatopic variations, such as Italian regional accents, since this is an aspect readily available to Italian students. In the context of other languages, the same concept could be however shown by different means, such as the influence of jargon or minority languages, or simply differences in accuracy depending on, age or other socio-demographic and socio-cultural variables. 

\paragraph{Part~3}
The next part of the activity (Section~\ref{sec:read}) is the most easily reproducible in a different language as it only exploits statistical co-occurrences as a cue for linguistic structure. The only pre-requisite here is the availability of a sufficiently standardized tokenization for the language of interest. As described above, we masked the language through a transliteration system: this was achieved either substituting words with sequences of symbols, or with non-words. In the case of non-words, these were generated by manually defining a series of phonotactic constraints for Italian, which should be adapted to the target language.

The main message to be conveyed through this activity is that language is a complex system; we did this disentangling semantics from the purely symbolic tier, which is the one computers most commonly manipulate.

\paragraph{Part~4}
The following part of the activity (Section~\ref{sec:grammar}) focuses on the expressive power residing in the definition of auxiliary categories as descriptors of linguistic evidence. We achieved this by simplifying a constituent-based annotation for our corpus: having continuous constituents easily allowed for the physical implementation of re-writing rules (i.e., participants had some physical constituents and tokens that could be used to simulate the generation process). We built categories in order to extract from the text a simple \textit{regular grammar}, and we were especially careful about the fact that both the Italian and the English corpus showed a similar structure and therefore complexity level. More specifically, we restricted to five types of higher-level categories that acted as phrases (i.e., sentence, noun phrase, verb phrase, prepositional phrase, subordinate clause): this is of course a huge simplification and, for the sake of the activity, we overlooked some relevant linguistic phenomena. 

We reckon that this approach might not be portable to languages that exhibit a flexible word order, so alternative solutions should be sought.

\paragraph{Part~5}
The last section of the activity was dedicated to a discussion on the presented methods for language generation.
After that and depending on the audience, we presented some options to pursue studies in Computational Linguistics in Italy, which would of course have to be adapted to the target social context. Lay publications concerned with Computational Linguistics are also unfortunately not very common in Italy, therefore we took the opportunity to provide participants with some suggestions for further readings \citep{masini2017tutto}.

\section{Looking back and ahead}

In the previous sections we described an interactive workshop designed to illustrate the basic principles of Computational Linguistics to young Italian students. It is the first activity of its kind to be designed by the Italian Association for Computational Linguistics and among the first dissemination activities in Italy for Computational Linguistics directed to young students.

The activity had the broad aim of increasing awareness towards applications based on language technology, and introduce students to the study of language as a scientific discipline.

We run the activity in both face-to-face and, due to COVID-19, online form: generally speaking, we received enthusiastic feedback both from younger participants and from the more general public.
We adapted the activity to a variety of formats and time-slots, ranging from 30 to 90 minutes: the amount of time required to approach the game and get acquainted with the concepts is of course variable, and depends on the participants' background and on the level of engagement that is expected of them. 
Generally speaking 45 minutes are enough for a presentation including some interaction with the audience, especially in the online setting, but at least 90 minutes are needed for the participants to fully experiment with the hands-on activity.

We want to specifically stress how time is a central ingredient in the activity. While the game-related aspects remained engaging and fun even in the shortened, online versions, in order to fully grasp the mechanisms underlying the presented algorithms longer sessions would be needed. 
In fact, we often felt that more time would be beneficial for a deeper discussion about language as an object of study itself, and about language as data on which theories can be built.
In particular, shorter time-slots or less guided activities enhance the risk of participants approaching the challenge as a puzzle that they can solve regardless of linguistic knowledge. This is because the approach to language we are presenting is entirely new to our audience, and not only to the younger students. 

Although we did not have a formal system in place to collect systematic feedback, the overall response across venues and conditions has been extremely positive.\footnote{Following one reviewer's suggestion, we have implemented such a system for our latest presentation at the Science Web Festival 2021 as a Google form that we circulated among participants at the end of the presentation.} 
Curiosity and engagement of participants remained high both onsite and online, and we received many questions on several aspects of Natural Language Processing and neural networks, as well as concerning its role in Artificial Intelligence at large.

The participants' enthusiastic questions gave us many ideas for future dissemination activities. In fact, the technological world is advancing fast and we firmly believe that it is necessary to spread more awareness on the inner workings of AI-based technologies, to develop a society-level conscience to approach them in a critical way.

The activity described in this paper was targeted at middle to high-school students as well as the general public. It would be interesting to engage with a younger audience as well, as communicating the study and (computational) modelling of language to them would raise awareness towards language studies as a scientific discipline.

 For our activity, we took inspiration from one of the problems proposed in \newcite{radev2013puzzles}. Puzzles such as the ones presented at the Computational Linguistics Competitions are a great way to introduce both important challenges and the methodology to solve them, as they stimulate students to investigate linguistic aspects in a bottom-up fashion. Organizing the competition in Italy would represent a bigger project for our association, to be addressed in the coming years.

\section*{Acknowledgements}
The workshop has been developed with help and support from many parties. We would like to thank them all here. First and foremost, all the participants that enthusiastically played along and made the activity a success. Along with them, four tutors and science animators helped us greatly in putting our ideas into practice. We are also grateful to BergamoScienza, Festival della Scienza, Science Web Festival for hosting the activity during the festivals, to ILC-CNR ``Antonio Zampolli" and ColingLab  (University of Pisa) for hosting us during the European Researchers' Night, and to the Scuola Internazionale Superiore di Studi Avanzati (SISSA) and ITI Tullio Buzzi. 
We are also grateful to Dr. Mirko Lai, who has collaborated on the development of the web interface for the online versions of our activity. We would like to thank AILC (Italian Association of Computational Linguistics) for encouraging us to design the activity and supporting us throughout the process. Last but not least, we thank the true hero of our workshop: GingerCat\footnote{\url{https://icons8.com/illustrations/style--ginger-cat-1}}.


\bibliography{anthology,custom}
\bibliographystyle{acl_natbib}

\end{document}